\documentclass[11pt]{article}

\usepackage[final]{acl}

\usepackage{times}
\usepackage{latexsym}

\usepackage[T1]{fontenc}
\usepackage[utf8]{inputenc}
\usepackage[most]{tcolorbox}
\usepackage{booktabs}
\usepackage{tabularx}
\usepackage{enumitem}
\usepackage{multirow}
\usepackage[table]{xcolor}
\definecolor{lightgray}{gray}{0.93}
\usepackage{microtype}
\usepackage{subcaption}
\usepackage{float}
\usepackage{xcolor}

\usepackage{xspace}
\newcommand{\ourmethod}{\textsc{PRISM}\xspace}

\newcommand{\CoT}{\textit{CoT}\xspace}
\newcommand{\van}{\textit{Vanilla}\xspace}

\newcommand{\socialbench}{\texttt{Social-Persona}\xspace}
\newcommand{\easypersona}{\texttt{Big5-Persona-EASY}\xspace}
\newcommand{\hardpersona}{\texttt{Big5-Persona-HARD}\xspace}

\usepackage{inconsolata}
\usepackage{graphicx}

\title{Do LLMs Understand Personality? Rethinking Persona Fidelity Evaluation through Structured Behavioral Inference}

\author{
\textbf{Mengfan Li\textsuperscript{1}\thanks{Work was done during a visit at SMU.}},
\textbf{Zesheng Wei\textsuperscript{2}},
\textbf{Xuanhua Shi\textsuperscript{1}\thanks{Corresponding author.}},
\textbf{Yang Deng\textsuperscript{2}} \\
\textsuperscript{1}National Engineering Research Center for Big Data Technology and System, \\
Services Computing Technology and System Lab, Cluster and Grid Computing Lab, \\
School of Computer Science and Technology, Huazhong University of Science and Technology \\
\textsuperscript{2}Singapore Management University \\
\texttt{\{limf, xhshi\}@hust.edu.cn, zswei66bx@gmail.com, ydeng@smu.edu.sg}
}

\begin{document}
\maketitle
\begin{abstract}
As large language models are increasingly deployed to simulate diverse human characters, ensuring \emph{persona fidelity}, defined as the extent to which an agent's behavior consistently reflects the psychological and stylistic characteristics of a target persona, has become a critical requirement. However, existing evaluation paradigms primarily rely on either holistic LLM-based judges, which are prone to ``holistic appraisal hallucination'', or static psychometric inventories, which fail to capture the context-dependent fidelity required in dynamic dialogue.  
To address these limitations, we propose \ourmethod (\textbf{P}ersona \textbf{R}easoning with \textbf{I}nverse \textbf{S}FL-based \textbf{M}odeling), a psycholinguistically grounded framework that reformulates persona fidelity evaluation as a structured inverse inference task. 
Inspired by Systemic Functional Linguistics (SFL), \ourmethod decomposes persona fidelity into three functional dimensions: \textit{Task Framing}, \textit{Interpersonal Stance}, and \textit{Linguistic Style}. 
It estimates dimension-specific evidence over a persona-conditioned label space and aggregates these signals into an interpretable and auditable evaluation process. Experiments show that \ourmethod yields more accurate and stable judgements than traditional holistic judging, providing a more reliable framework for persona fidelity evaluation. 
\end{abstract}

\section{Introduction}

Recent advances in Large Language Models (LLMs) have enabled increasingly sophisticated role-playing agents that can simulate diverse personas and social identities \cite{tu2024charactereval, li2025big5}. 
As these agents are increasingly deployed in immersive and interactive environments, ensuring their \emph{consistency} with assigned characters has emerged as a crucial desideratum \cite{ji2025enhancing, wang2024rolellm, bhandari2025can}. A compelling role-playing agent should not only generate coherent responses that remain consistent with persona-related knowledge, but also maintain stable and recognizable personality traits and behavioral styles
throughout interaction \cite{wu2025raiden, li2026costom}. This requirement is commonly referred to as \textbf{persona fidelity}: the extent to which a model’s behavior consistently reflects the psychological and stylistic characteristics of a target persona \cite{shin2025spotting, wang2024incharacter}. 

\begin{figure}[t]
\setlength{\abovecaptionskip}{3pt}   
    \setlength{\belowcaptionskip}{0pt}
    \centering
    \includegraphics[width=1.0\linewidth]{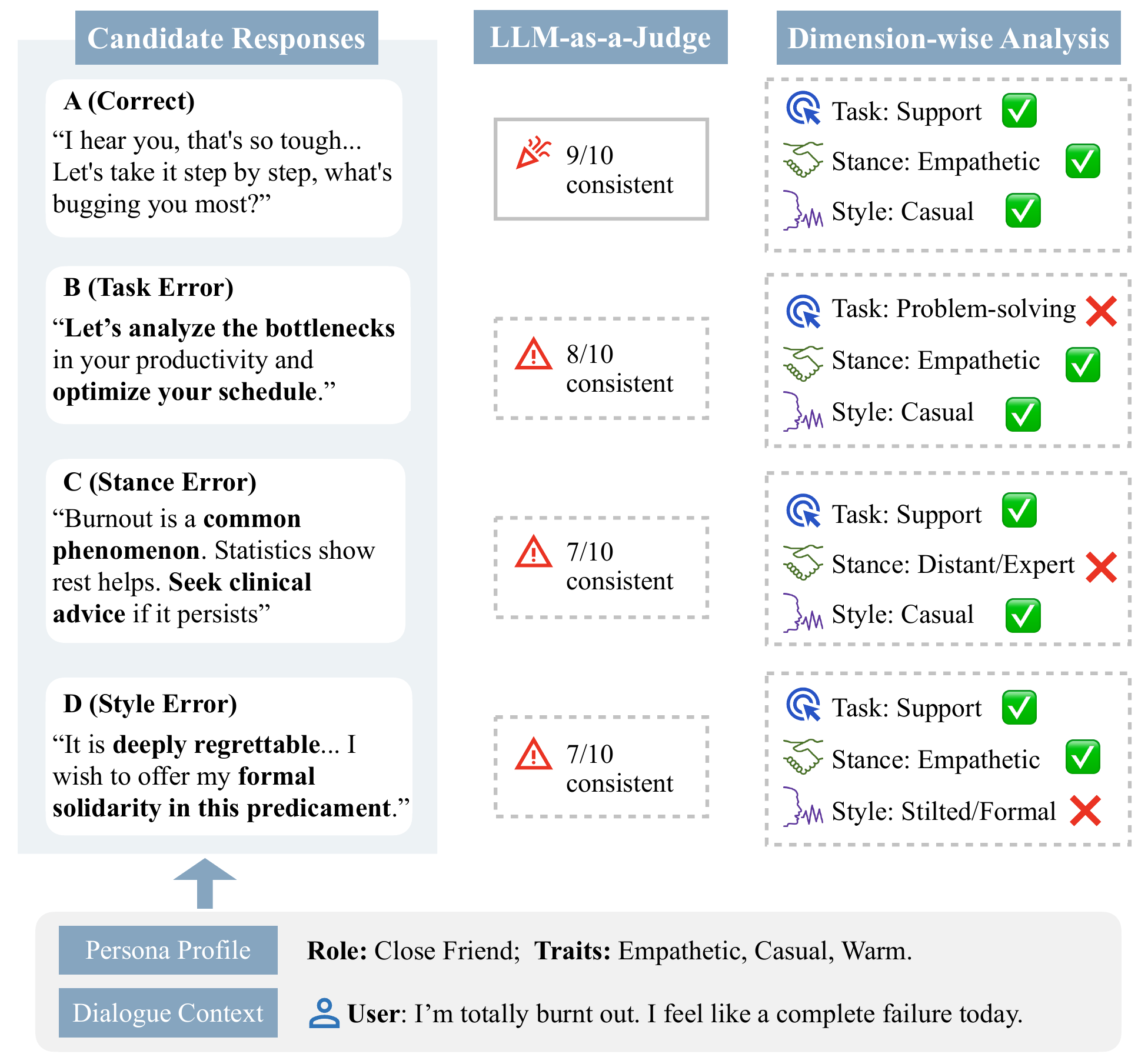}
    \caption{Holistic vs. Dimension-wise evaluation of persona fidelity. Holistic judges are often misled by surface-level fluency (B-D), whereas our dimension-wise analysis provides interpretable and diagnostic evidence for persona (mis)alignment across three functional dimensions.}
    \label{motivation}
    \vspace{-5mm}
\end{figure}

Despite its importance, reliably evaluating persona fidelity remains a significant challenge \cite{jiang2024personallm, yoon2024evaluating, ji2025enhancing}. 
Importantly, persona fidelity differs fundamentally from conventional notions of factual consistency in personalized dialogue systems \cite{zhang2018personalizing, shao2023character, mazare2018training}, which focus on whether a model can accurately recall or reproduce user-specific facts, such as demographic attributes, preferences, or biographical information. In contrast, persona fidelity concerns whether the model \emph{behaves} in a manner aligned with the underlying personality and behavioral style of the assigned character \cite{jiang2023evaluating}.
A response may correctly mention persona-related facts while still deviating from the target persona in nuanced psychological or stylistic ways. 
Such discrepancies are rarely captured by surface-level semantic similarity or simple factual matching. 
Consequently, robust evaluation hinges on the ability to distinguish truly ``in character'' responses from plausible but behaviorally misaligned alternatives.  

Current evaluation paradigms for persona fidelity 
follow two methodological categories. 
The most prevalent is \textbf{LLM-as-a-judge}, in which an evaluator model directly assigns a holistic consistency score to a response \cite{tu2024charactereval, wang2024rolellm, zhou2024sotopia}. 
While scalable, this approach is often nontransparent and prone to ``holistic appraisal hallucination'', where judges overrate fluent but out-of-character responses \cite{wu2025style, shin2025spotting, DBLP:conf/iclr/WangYYZYW0J000024, wang2024large, li2026rectom}. 
Another paradigm, \textbf{psychometric probing}, assesses agents through standardized personality inventories (\textit{e.g.}, Big Five or MBTI) \cite{wang2024incharacter, jiang2024personallm}. While effective for trait-level analysis, these methods typically rely on static interviews, thereby failing 
to capture the fine-grained and context-dependent fidelity required in spontaneous and dynamic dialogues.

In this work, we argue that persona fidelity should be evaluated as a structured, multidimensional consistency problem. 
Drawing inspiration from \emph{Systemic Functional Linguistics (SFL)} \cite{halliday2013halliday}, we decompose consistency into three functional dimensions: \textbf{Task Framing}, \textbf{Interpersonal Stance}, and \textbf{Linguistic Style}. 
From this perspective, persona is materialized not only through \emph{what} an agent says, but also through \emph{how} it frames goals, negotiates interpersonal relationships, and adopts characteristic linguistic patterns.
As shown in Figure \ref{motivation}, while holistic judges are frequently misled by surface-level helpfulness, our dimension-wise decomposition enables 
fine-grained identification of why and how a response deviates from the target persona.

Based on this framework, we propose \ourmethod (\textbf{P}ersona \textbf{R}easoning with \textbf{I}nverse \textbf{S}FL-based \textbf{M}odeling), a structured evaluation framework for persona fidelity. 
Unlike holistic judges, \ourmethod reformulates evaluation as an \emph{inverse structured inference task}: given a response and context, the evaluator infers dimension-specific evidence and checks its alignment with the target persona. Specifically, \ourmethod estimates a posterior distribution over a profile-conditioned label space (\textit{Aligned}, \textit{Indeterminate}, or \textit{Contradictory}) for each SFL-based dimension. By aggregating these fine-grained signals into an \emph{Inverse Persona Evidence}, \ourmethod provides 
a more interpretable, psycholinguistically grounded, and auditable evaluation process for persona fidelity.

Given the lack of dedicated benchmarks for evaluating persona fidelity evaluation frameworks, we construct three diagnostic benchmarks, namely \easypersona, \hardpersona, and \socialbench, based on existing persona-consistent dialogue corpora \cite{li2025big5, chen2024socialbench}.
To rigorously assess evaluator reliability, we introduce controlled perturbation strategies to generate \emph{hard negative} responses that remain contextually 
plausible while subtly violating the target persona’s behavioral or linguistic style.

Experimental results show that \ourmethod consistently outperforms traditional holistic judges. 
Furthermore, our analysis reveals that the functional decomposition effectively mitigates the ``holistic appraisal hallucination'' and exhibits superior stability across varying evaluator backbones and scoring rubrics, establishing \ourmethod as a reliable and interpretable framework for persona fidelity assessment.

Our contributions are threefold:
\begin{itemize}[leftmargin=*]
    \item 
    \textbf{Psycholinguistically-grounded Formalization}: We formalize persona fidelity as a structured, multidimensional behavioral consistency problem. Inspired by \emph{Systemic Functional Linguistics}, we decompose persona-relevant behavior into three functional dimensions: task framing, interpersonal stance, and linguistic style.
    \item 
    \textbf{Evaluation Framework}: We propose \ourmethod, a structured evaluation framework that reformulates persona evaluation as an \emph{inverse structured inference} task. By estimating dimension-specific posterior distributions, \ourmethod provides an interpretable and auditable evaluation process.
    \item \textbf{Benchmarks and Validation}: 
    We curate three diagnostic benchmarks with contextually plausible hard negatives for evaluating persona fidelity assessment methods.
    Extensive analyses show that \ourmethod consistently outperforms holistic judges in reliability and robustness~\footnote{Code and data: \url{https://github.com/CGCL-codes/prism-persona}}.
\end{itemize}

\section{Related Work}
\paragraph{Personalization for Role-playing Agents} 
Recent advances in Large Language Models (LLMs) have enabled increasingly sophisticated role-playing agents capable of embodying diverse personas and social identities \cite{tois22-persona,acl23f-persona,emnlp24-usersim,zhou2024characterglm, shin2025spotting, peng2026rethinking, yang2025consistent, qiu2026psyclient, zhu2025llm, chen2025socialsim}. 
The efficacy of role-playing agents is intrinsically tied to \textbf{personalization}, which aims to transform generic LLMs into distinct, recognizable personas \cite{li2025big5, DBLP:conf/eacl/AraujoHMSR26}. 
Prior work has studied personalization generation from two related perspectives: \textbf{factually-consistent} and \textbf{personality-grounded}. 

\emph{Factually-consistent generation} emphasizes accurate recall of persona-related information, often through retrieval-augmented generation \cite{emnlp23f-persona,wang2024rolellm} or memory mechanisms \cite{xu2022long, he2025madial,naacl25-ldagent} to preserve biographical details such as age, occupation, and experiences \cite{shao2023character}. 
In contrast, \emph{personality-grounded generation} seeks to induce stable psychological traits and behavioral styles through psychometric prompting (\textit{e.g.}, Big Five or MBTI) \cite{de2000big, jiang2023evaluating,emnlp25-persona}, steering \cite{acl26f-persona}, or character-specific fine-tuning \cite{wang2024rolellm, li2023chatharuhi}.

Despite these advances, existing evaluation frameworks primarily focus on factual consistency \cite{tan2025personabench, he2025chinese, chen2025towards}, assessing whether agents can correctly reproduce persona-related facts. However, factual consistency alone is insufficient for high-quality role-playing: an agent may accurately recall persona information while still failing to exhibit the intended personality traits or behavior styles.
Our work addresses this gap by shifting evaluation from \emph{``what the agent knows''} (fact) to \emph{``how the agent behaves''} (persona fidelity).


\paragraph{Methodologies for Persona Fidelity Evaluation}
Existing approaches for persona fidelity evaluation mainly follow two paradigms: \textbf{holistic appraisal} \cite{wang2025opencharacter} and \textbf{psychological probing} \cite{ye2025large,wang2024incharacter}. 
\emph{Holistic Appraisal} typically adopts an LLM-as-a-judge framework, where an evaluator model assigns a single consistency score to generated responses \cite{jun2025exploring, zhou2024sotopia, feng2025reasoning}. 
While scalable, this approach is susceptible to ``holistic appraisal hallucination'' \cite{wu2025style, shu2024you}, where judges are frequently misled by surface-level fluency or the ``helpfulness bias'' \cite{wu2025style, zheng2023judging}. This often results in rating polite or informative responses favorably while overlooking subtle persona violations. 
\emph{Psychological probing} assesses persona through standardized psychological inventories, such as the Big Five Inventory \cite{jiang2023evaluating, bhandari2025can, jiang2024personallm} or MBTI \cite{tu2023characterchat, tu2024charactereval}. 
Although effective for trait-level analysis, these methods are typically based on static questionnaires or decontextualized interviews \cite{wang2024incharacter}, limiting their ability to capture fine-grained and context-dependent persona fidelity in dynamic dialogue.

In contrast to prior work, we formulate persona fidelity evaluation as a structured and interpretable consistency problem. Our framework decomposes persona-consistent behavior into multiple functional dimensions, enabling fine-grained diagnosis of subtle behavioral deviations beyond single-score holistic judgments.

\begin{figure*}[t]
\setlength{\abovecaptionskip}{3pt}   
    \setlength{\belowcaptionskip}{0pt}
    \centering
    \includegraphics[width=1.0\linewidth]{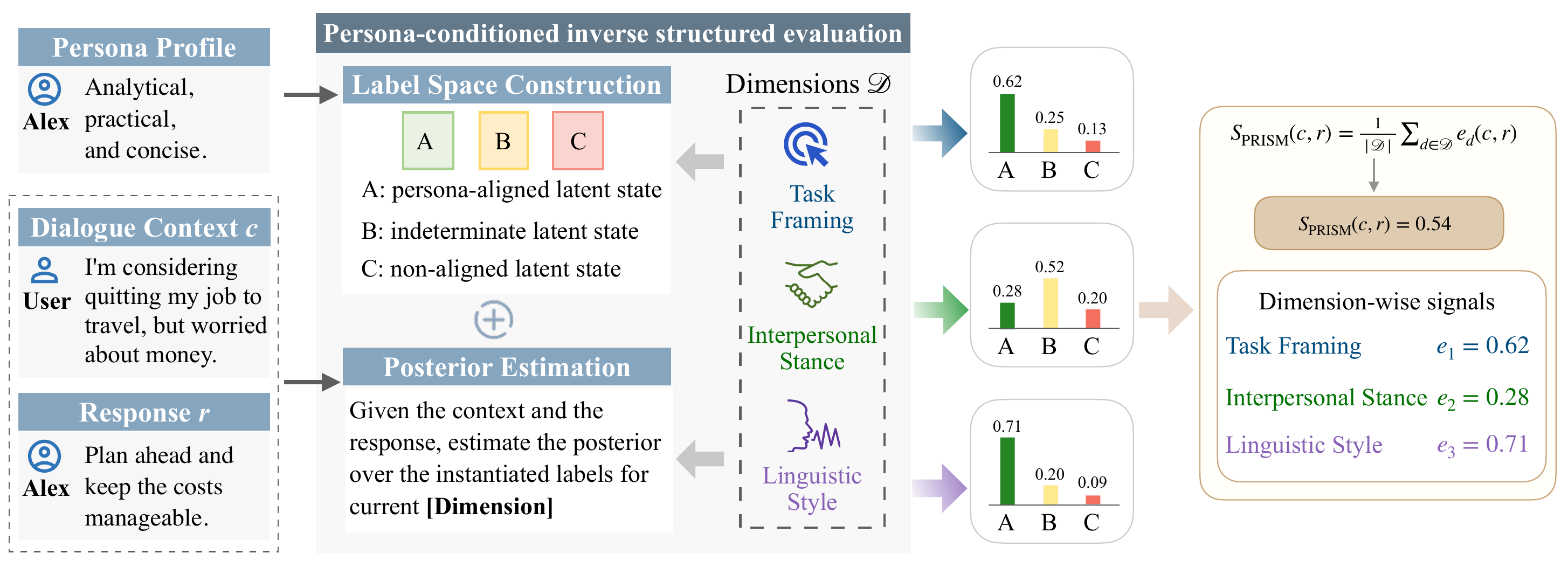}
    \caption{Overview of the \ourmethod framework. \ourmethod constructs persona-conditioned latent spaces across three functional dimensions and performs inverse posterior estimation over the instantiated labels. The dimension-level signals ($e_d$) are then aggregated into a diagnostic persona fidelity assessment.}
    \label{framework}
    \vspace{-3mm}
\end{figure*}

\paragraph{Systemic Functional Linguistics} 
Systemic Functional Linguistics (SFL) views language as a resource for meaning-making in social context \cite{halliday2013halliday, eggins2004introduction, matthiessen2023systemic}. 
A central perspective in SFL is that language simultaneously realizes multiple metafunctions: the ideational metafunction for representing experiences and events, the interpersonal metafunction for enacting social relations, and the textual metafunction for organizing meanings in discourse \cite{thompson2019cambridge}. 

This functional perspective is particularly relevant to persona fidelity because a response may be contextually appropriate while still differing from the target persona in how it construes the interaction, relates to the interlocutor, or expresses itself linguistically \cite{bucholtz2005identity, agha2006language}. 
Motivated by this perspective, PRISM organizes persona-relevant evidence along three operational dimensions: \textit{Task Framing}, which captures the activity orientation or communicative goal foregrounded by the response \cite{halliday2013halliday}; \textit{Interpersonal Stance}, which captures the relational position enacted toward the interlocutor \cite{jaffe2009stance}; and \textit{Linguistic Style}, which captures characteristic patterns in how the response is linguistically expressed \cite{coupland2007style}. 
These dimensions provide interpretable, complementary views of persona realization.

\section{\ourmethod Evaluation Framework}
Instead of directly asking whether a response is consistent with a target persona, we formulate persona fidelity evaluation as a \emph{persona-conditioned inverse structured evaluation} problem.
Following the theory of \emph{Systemic Functional Linguistics} \cite{halliday2013halliday}, \ourmethod decomposes persona fidelity into three interpretable and psycholinguistically-grounded dimensions: \textit{task framing}, \textit{interpersonal stance}, and \textit{linguistic style}.
Let $\mathcal{D} = \{d_1, d_2, d_3\}$ denote the three dimensions. 
For each dimension, \ourmethod performs inverse inference over the dialogue context and candidate response to estimate how strongly the response expresses the persona-aligned latent behavioral state. Concretely, it proceeds in two main steps, as shown in Figure \ref{framework}. 

\paragraph{Persona-Conditioned Label Space Construction} 
For each dimension $d \in \mathcal{D} $, \ourmethod defines a dimension-specific, persona-conditioned label space \( \mathcal{Y}_d= \{A, B, C\} \). 
Here, $A$ denotes the persona-aligned latent state for dimension $d$, $B$ denotes an indeterminate or mixed state, and $C$ denotes an opposite or non-aligned state. 
The semantic interpretations of $A$, $B$, $C$ are defined separately for each dimension and relative to the target persona. 
Accordingly, these labels represent dimension-level latent states rather than instance-level positive/negative labels, and a response may remain aligned on some dimensions while deviating on others. 
Figure \ref{label_space_case} illustrates one concrete label-space instantiation for the \textit{Interpersonal Stance} dimension under a target persona characterized by \textit{High Agreeableness}. 
Additional 
dataset-specific cases and construction details are provided in Appendix~\ref{label_space_appendix}.

\begin{figure}[h]
\setlength{\abovecaptionskip}{3pt}   
    \setlength{\belowcaptionskip}{0pt}
    \centering
    \begin{tcolorbox}[
        enhanced,
        sharp corners,
        boxrule=0.5pt,
        colback=gray!3,
        colframe=gray!70,
        fonttitle=\bfseries\small,
        title=Instantiated Label Space Example,
        fontupper=\small,
        width=0.98\columnwidth,
        left=8pt,
        right=8pt,
        top=6pt,
        bottom=6pt
    ]
\textbf{Options:}

\vspace{0.4em}
\noindent \textbf{A}\hspace{0.8em}warm, accommodating, and harmony-managing interpersonal stance

\vspace{0.35em}
\noindent \textbf{B}\hspace{0.8em}weakly marked, mixed, flat, generic, or insufficiently diagnostic interpersonal stance

\vspace{0.35em}
\noindent \textbf{C}\hspace{0.8em}blunt, less accommodating, or hard-edged interpersonal stance
    \end{tcolorbox}
    \caption{Instantiated label space for the \textit{Interpersonal Stance} dimension under a target persona characterized by \textit{High Agreeableness}.}
 
    \label{label_space_case}
    \vspace{-3mm}
\end{figure}

\paragraph{Inverse Posterior Estimation} 
Given a dialogue context $c$ and a candidate response $r$, \ourmethod constructs a \emph{dimension-specific inverse prompt} and estimates the model's conditional support for each label in $\mathcal{Y}_d$. 
These scores are normalized over the restricted label space to obtain a posterior-like distribution:
\begin{equation}\small
    q_d(y \mid c, r) = \frac{\exp(s_d(y \mid c, r))}{\sum_{y' \in \mathcal{Y}_d} \exp(s_d(y' \mid c, r))}, \quad y \in \mathcal{Y}_d,
\end{equation}
where $s_d(y \mid c, r)$ denotes the model's conditional log-score assigned to the label completion corresponding to $y$ under the inverse prompt for dimension $d$. 
Unlike holistic free-form judging, \ourmethod performs evaluation by scoring a restricted set of structured label completions and normalizing their relative support. This design reduces ambiguity in evaluator generation and constrains the evaluation process to explicitly defined behavioral states.
To reduce label-position bias, we randomly permute the displayed label order for each prompt and map model outputs back to the canonical \textit{aligned} / \textit{neutral} / \textit{non-aligned} label space before scoring. 

We use the aligned-state probability as the dimension-level consistency signal:
\begin{equation}\small
e_d (c, r) = q_d (A \mid c, r).
\end{equation}
This score quantifies how strongly the response expresses the persona-consistent latent state along dimension $d$. 
The final persona fidelity score is estimated by averaging the inverse evidence across dimensions: 
\begin{equation}\small
S_{\mathrm{\ourmethod}}(c,r) = \frac{1}{|\mathcal{D}|}\sum\nolimits_{d \in \mathcal{D}} e_d(c, r).
\end{equation}
This design yields two advantages. 
First, it turns persona evaluation from an opaque end-to-end rating problem into a structured set of interpretable sub-decisions. 
Second, it preserves diagnostic granularity: beyond the final score $S_{\mathrm{\ourmethod}}(c, r)$, the individual dimension scores $\{e_d\}_{d \in \mathcal{D}}$ reveal which aspect of persona realization is aligned or misaligned in the response.

\section{Experimental Details}
\subsection{Dataset}
Given the absence of available benchmarks for evaluating persona fidelity, we construct three evaluation datasets from existing personalized generation benchmarks: \texttt{Big5-Persona-EASY}, \texttt{Big5-Persona-HARD}, and \texttt{Social-Persona}.

The Big5-based benchmarks are derived from Big5-CHAT \cite{li2025big5}, which provides dialogue triplets: \((p, c, r)\), where $p$ denotes a target profile (\textit{e.g.}, \textit{High Agreeableness}), $c$ is the dialogue context, and $r$ is a persona-consistent response. 
We construct ``hard negatives'' by minimally disturbing the alignment within a triplet while keeping other elements fixed: 
(1) \textbf{\easypersona}. We maintain the context $c$ and response $r$ but substitute $p$ with its direct opposite profile $p'$ within the same personality dimension (\textit{e.g.}, replacing \textit{High Agreeableness} with \textit{Low Agreeableness}). This setup evaluates the model's sensitivity to directional tendencies of a specific trait. 
(2) \textbf{\hardpersona}: To simulate subtler misalignments, we construct ``near-miss'' negatives by cross-matching traits across different dimensions: either (i) \((p', c, r)\) where $p'$ belongs to a different trait dimension entirely (\textit{e.g.}, swapping \textit{High Extraversion} for \textit{High Agreeableness}), or (ii) \((p ,c, r')\), where $r'$ is a response generated for a different trait within the same scenario. 
These cases require the model to distinguish between fine-grained behavioral realizations that share surface-level similarities, as the responses remain contextually plausible yet violate the specific behavioral constraints of the target persona. 

\textbf{\socialbench} derives from the role-style subset of SocialBench \cite{chen2024socialbench}. We convert the original multiple-choice format into a point-wise evaluation setting by pairing the target profile and context with each candidate response independently to form multiple \((p, c, r)\) triplets.


Table \ref{statistics} presents the dataset statistics. In our experiments, we randomly sample a subset of the Big5-based benchmarks, comprising 2,000 instances for \texttt{Big5-Persona-EASY} and 3,000 for \texttt{Big5-Persona-HARD}. Examples of the dataset construction are detailed in Appendix~\ref{dataset_construction}. We further validate the reformulated evaluation instances through a benchmark-level human study; details and results are provided in Appendix~\ref{human}.

\begin{table}[t]
\setlength{\abovecaptionskip}{3pt}   
    \setlength{\belowcaptionskip}{0pt}
\centering
\small
\begin{tabular}{lccc}
\toprule
Dataset & Size & Pos:Neg  \\
\midrule
Big5-Persona-EASY & 200K & 1:1  \\
Big5-Persona-HARD & 300K & 1:2  \\
Social-Persona & 2.9K & 1:3 \\
\bottomrule
\end{tabular}
\caption{Statistics of the persona fidelity benchmarks.}
\label{statistics}
\vspace{-3mm}
\end{table}
\begin{table*}[t]
\centering
\small
\setlength{\tabcolsep}{4.5pt}
\renewcommand{\arraystretch}{1}

\begin{tabular}{llccccccccc}
\toprule
\multirow{2}{*}{Model} & \multirow{2}{*}{Method}
& \multicolumn{3}{c}{Social-Persona}
& \multicolumn{3}{c}{Big5-Persona-EASY}
& \multicolumn{3}{c}{Big5-Persona-HARD} \\
\cmidrule(lr){3-5} \cmidrule(lr){6-8} \cmidrule(lr){9-11}
& & AUC & P-AUC & G-Acc
  & AUC & P-AUC & G-Acc
  & AUC & P-AUC & G-Acc \\
\midrule

\multicolumn{11}{c}{\textit{Specialized Evaluation Models}} \\
\midrule
Selene-Mini & \multicolumn{1}{c}{--}
& 87.50 & 88.31 & 52.29
& 84.90 & 82.75 & 74.40
& 63.06 & 66.02 & 23.70 \\

PandaLM & \multicolumn{1}{c}{--}
& -- & 48.28 & 10.13
& -- & 48.00 & 48.00
& -- & 48.45 & 23.60 \\

AlignScore & \multicolumn{1}{c}{--}
& 56.77 & 57.20 & 32.56
& 53.80 & 53.60 & 53.60
& 52.72 & 55.75 & 31.90 \\
\midrule

\multicolumn{11}{c}{\textit{Open-source LLMs}} \\
\midrule
\multirow{3}{*}{Qwen}
& Vanilla      & 83.26 & 84.34 & 49.32 & 86.05 & 85.85 & 77.80 & 68.60 & 71.70 & 36.50 \\
& +CoT         & \textbf{84.23} & 85.09 & 50.00 & 83.06 & 82.45 & 72.00 & 65.94 & 68.77 & 28.40 \\
\rowcolor{gray!12}
& \ourmethod   & 84.15 & \textbf{91.08} & \textbf{78.78}
               & \textbf{96.29} & \textbf{97.60} & \textbf{97.60}
               & \textbf{75.99} & \textbf{80.85} & \textbf{68.20} \\
\midrule

\multirow{3}{*}{Llama}
& Vanilla      & 73.26 & 73.28 & 18.78 & 72.76 & 70.85 & 55.20 & 58.15 & 58.32 & 3.90 \\
& +CoT         & 72.48 & 73.53 & 20.94 & 72.76 & 70.85 & 55.20 & 60.61 & 60.50 & 18.10 \\
\rowcolor{gray!12}
& \ourmethod   & \textbf{79.73} & \textbf{88.96} & \textbf{75.27}
               & \textbf{89.31} & \textbf{93.30} & \textbf{93.30}
               & \textbf{66.97} & \textbf{75.90} & \textbf{60.30} \\
\midrule

\multirow{3}{*}{Mistral}
& Vanilla      & 78.20 & 81.19 & 52.16 & 65.01 & 64.15 & 50.60 & 53.55 & 57.50 & 20.00 \\
& +CoT         & \textbf{83.85} & 86.53 & 59.32 & 67.89 & 65.75 & 57.60 & 56.00 & 57.75 & 19.00 \\
\rowcolor{gray!12}
& \ourmethod   & 78.90 & \textbf{86.84} & \textbf{70.40}
               & \textbf{85.58} & \textbf{85.90} & \textbf{85.90}
               & \textbf{64.23} & \textbf{71.67} & \textbf{54.90} \\
\midrule

\multicolumn{11}{c}{\textit{Closed-source LLMs}} \\
\midrule
\multirow{2}{*}{DeepSeek-V3.2}
& Vanilla      & \textbf{88.21} & \textbf{89.68} & \textbf{60.94}
               & 89.05 & \textbf{88.55} & \textbf{81.00}
               & 65.76 & 67.35 & 26.80 \\
& +CoT         & 86.21 & 87.36 & 52.56
               & \textbf{89.27} & 88.45 & 79.60
               & \textbf{67.93} & \textbf{69.55} & \textbf{28.80} \\
\midrule

\multirow{2}{*}{GPT-5.4}
& Vanilla      & 91.19 & 92.50 & 69.00
               & 98.96 & 98.50 & 97.00
               & 77.96 & 78.90 & 44.60 \\
& +CoT         & \textbf{92.64} & \textbf{93.56} & \textbf{73.00}
               & 98.96 & 98.50 & 97.00
               & \textbf{78.60} & \textbf{79.10} & \textbf{46.60} \\
\midrule

\multirow{2}{*}{Gemini-3-Flash}
& Vanilla      & 92.02 & 92.65 & 69.81
               & 97.48 & \textbf{97.44} & \textbf{94.89}
               & 76.22 & 76.98 & \textbf{40.88} \\
& +CoT         & \textbf{92.73} & \textbf{93.39} & \textbf{72.83}
               & \textbf{97.83} & 96.93 & 93.87
               & \textbf{76.52} & \textbf{77.18} & 40.68 \\
\bottomrule
\end{tabular}
\caption{Main results on three persona fidelity benchmarks. AUC measures overall ranking quality, P-AUC denotes Pair-AUC, and G-Acc denotes strict Group Accuracy. Dashes indicate inapplicable metrics; in particular, PandaLM is evaluated only in pairwise form and therefore does not admit AUC. Within each model family, the best result is boldfaced, and \ourmethod\ rows are highlighted in gray.}
\label{main_results}
\vspace{-3mm}
\end{table*}

\subsection{Models}
We evaluate persona fidelity using nine LLM-based evaluators. 
Our main open-source evaluator backbones are Qwen2.5~\cite{DBLP:journals/corr/abs-2412-15115}, Llama-3.1~\cite{grattafiori2024llama}, and Mistral~\cite{DBLP:journals/corr/abs-2310-06825}. We further include three stronger external evaluators: DeepSeek-V3.2~\cite{liu2025deepseek}, GPT-5.4 and Gemini-3-Flash, as reference judges for direct LLM-as-a-judge evaluation. 
In addition, we consider three specialized evaluation models. 
\textbf{Atla Selene Mini} \cite{DBLP:conf/eacl/AraujoHMSR26} is a state-of-the-art small language model-as-a-judge fine-tuned model for general-purpose evaluation. 
\textbf{PandaLM-7B-v1}~\cite{DBLP:conf/iclr/WangYYZYW0J000024} is a Llama-7B-based response-comparison judge, which we adapt to select the more profile-consistent response in each pair. 
\textbf{AlignScore-large}~\cite{zha2023alignscore} is a RoBERTa-large-based factual consistency evaluator, which we adapt by treating the serialized profile and dialogue context as the reference and candidate response as the claim.

\subsection{Evaluation Metrics}

We evaluate model performance using three ranking-based metrics: AUC, Pair-AUC (P-AUC), and Strict Group Accuracy (G-Acc). 
Let $\mathcal{G}$ denote the set of all contrastive groups. For each group $g \in \mathcal{G}$, let $r_g^{+}$ denote the persona-consistent response and let $\{r_{g, j}^{-}\}_{j=1}^{m_g}$ denote the set of $m_g$ negative responses under the same profile and dialogue context. 
Let $s(\cdot)$ be the consistency score assigned by a method. 
For any positive-negative pair, we define the comparison function

\begin{equation}\small
\phi(r_g^{+}, r_{g,j}^{-}) = 
\begin{cases}
    1, & s(r_g^{+}) > s(r_{g,j}^{-}), \\
    0.5, &s(r_g^{+}) = s(r_{g,j}^{-}) , \\
    0, & s(r_g^{+})<s(r_{g,j}^{-}).
\end{cases}
\end{equation}
 
\textbf{AUC} measures global ranking quality over all positive and negative responses. It is defined as the probability that a randomly sampled persona-consistent response receives a higher score than a randomly sampled profile-inconsistent response:
\begin{equation} \small
\mathrm{AUC} = 
\frac{1}{|\mathcal{P}|\,|\mathcal{N}|}
\sum\nolimits_{r^{+}\in\mathcal{P}}\sum\nolimits_{r^{-}\in\mathcal{N}}\phi(r^{+}, r^{-}),
\end{equation}
where $\mathcal{P} = \{r_g^{+} \mid g \in \mathcal{G}\}$, $\mathcal{N} = \{r_{g,j}^{-} \mid g \in \mathcal{G}, 1 \le j \le m_g\}$.

\textbf{Pair-AUC} measures whether the target-consistent response receives a higher score than its contrastive alternatives, by averaging all within-group positive-negative pairs:
\begin{equation}\small
\mathrm{Pair\mbox{-}AUC} = \frac{\sum_{g \in \mathcal{G}} \sum_{j=1}^{m_g} \phi(r_g^{+}, r_{g,j}^{-})}{\sum_{g \in \mathcal{G}} m_g}.
\end{equation} 

\textbf{Strict Group Accuracy} (G-Acc) measures whether the positive response is ranked above \emph{all} negative responses within the same group: 
\begin{equation}\small
\mathrm{G\mbox{-}ACC} = 
\frac{1}{|\mathcal{G}|}\sum_{g \in \mathcal{G}} \mathbf{1}\!\left[s(r_g^{+})>\!\max_{1 \leq j \leq m_g} s(r_{g,j}^{-}) \right],
\end{equation}
where $\mathbf{1}[\cdot]$ is the indicator function. This metric is stricter than Pair-AUC, since it requires the persona-consistent response to outrank every negative candidate in its contrastive group.

\begin{figure*}[t]
\setlength{\abovecaptionskip}{3pt}   
    \setlength{\belowcaptionskip}{0pt}
    \centering
    \includegraphics[width = 1.0\linewidth]{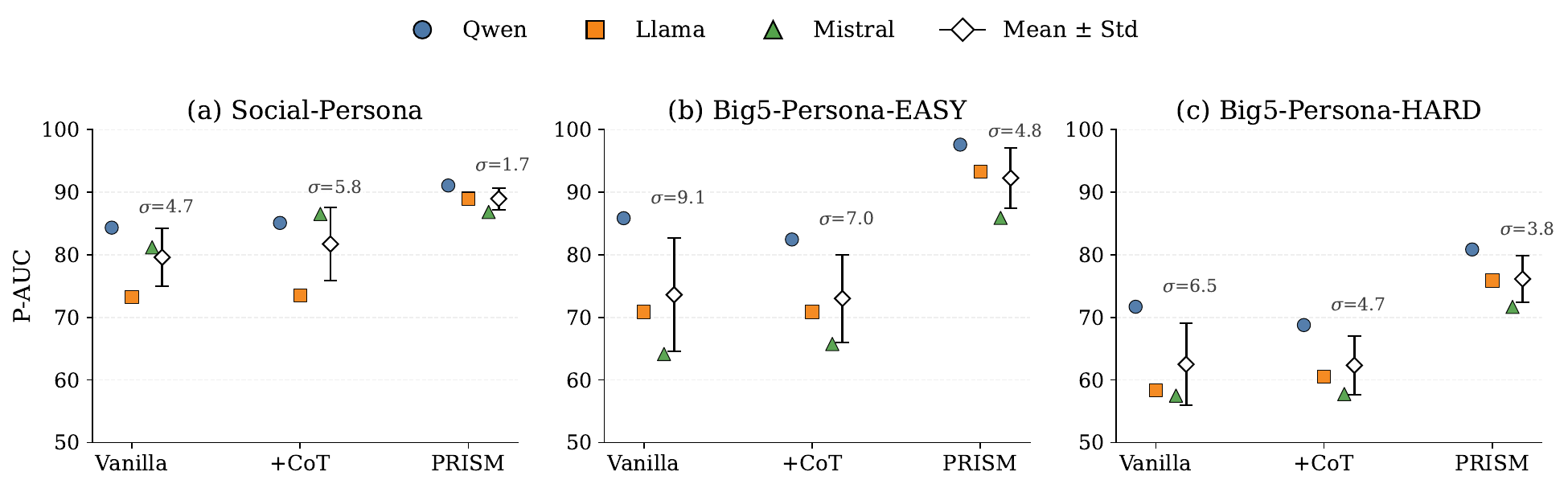}
    \caption{Backbone sensitivity on Pair-AUC across three LLM backbones. Points denote individual backbones and diamonds indicate the mean with standard deviation.}
    \label{backbone_sensitivity_pauc}
    \vspace{-3mm}
\end{figure*}

\begin{figure*}
\setlength{\abovecaptionskip}{3pt}
\setlength{\belowcaptionskip}{0pt}
    \centering
    \includegraphics[width=1.0 \linewidth]{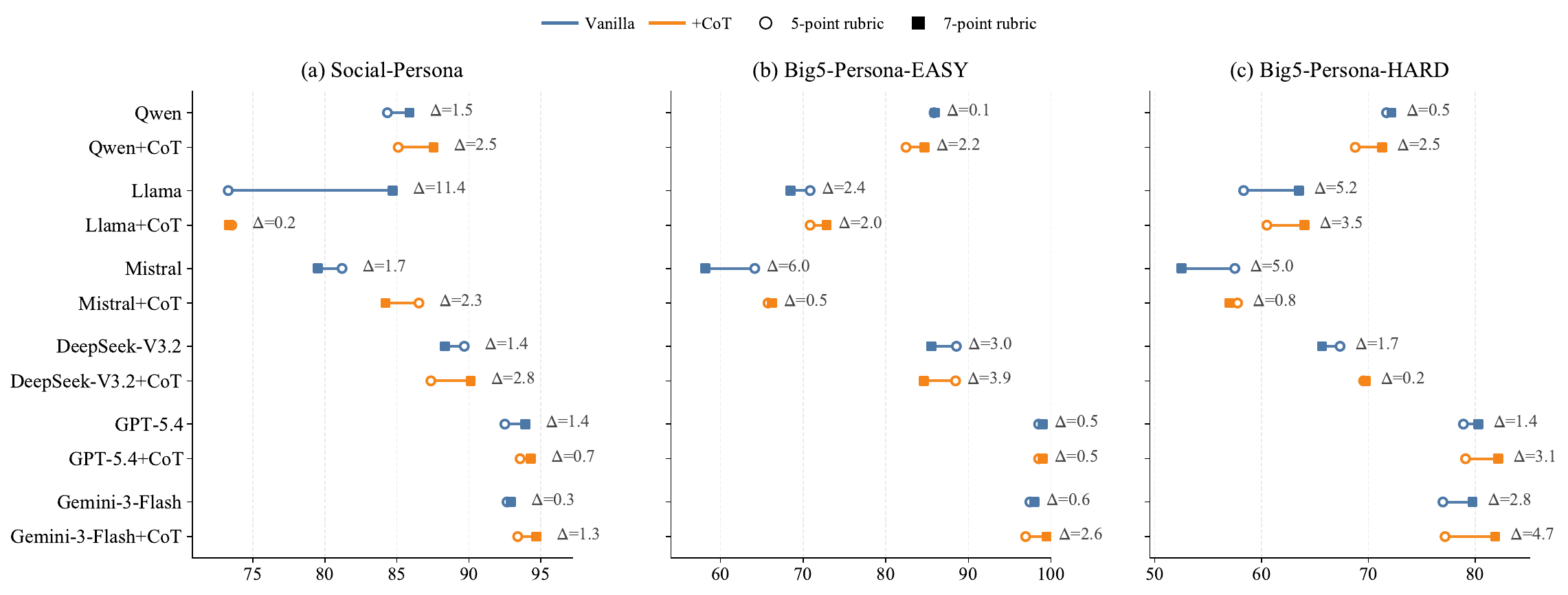}
    \caption{Rubric Sensitivity of direct LLM-as-a-Judge evaluation on Pair-AUC. Each segment connects the results from 5-point and 7-point rubrics for the same evaluator and prompting method. Longer segments indicate greater sensitivity to scoring granularity.}
    \label{rubric_sensitivity_pauc}
    \vspace{-3mm}
\end{figure*}

\begin{figure*}[t]
\setlength{\abovecaptionskip}{-10pt}   
    \setlength{\belowcaptionskip}{0pt}
    \centering
    \includegraphics[width=1.0\linewidth]{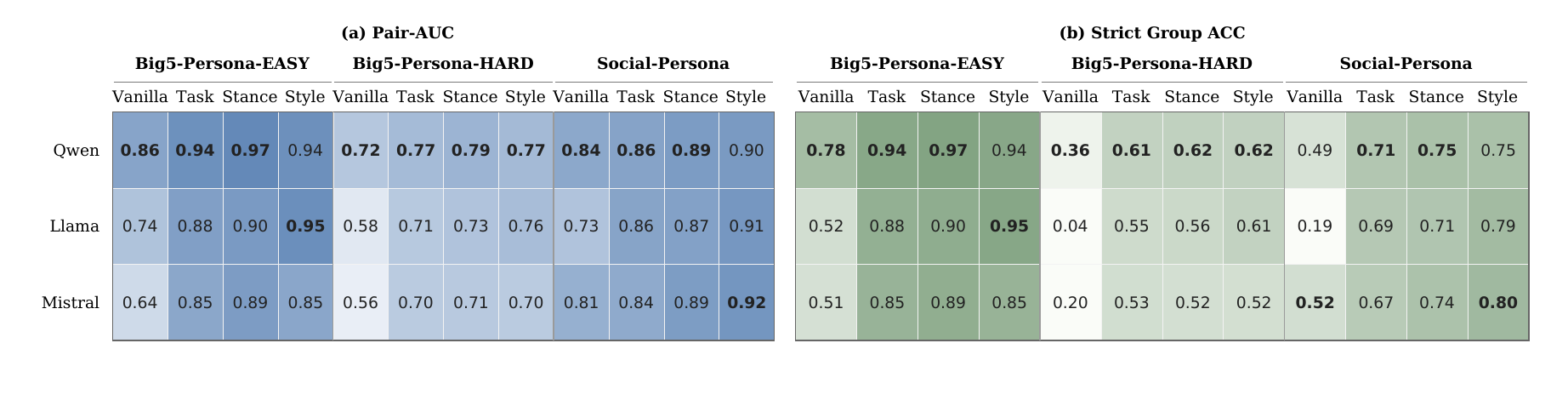}
    \caption{Dimension-wise diagnostic evaluation on three persona fidelity benchmarks. \van denotes holistic scoring, while ``Task'', ``Stance'', and ``Style'' denote single-dimension scoring based on task framing, interpersonal stance, and linguistic style. Darker cells indicate stronger performance.}
    \label{single_dimension}
    \vspace{-3mm}
\end{figure*}

\begin{figure*}[t]
\setlength{\abovecaptionskip}{-10pt}   
    \setlength{\belowcaptionskip}{0pt}
    \centering
\includegraphics[width=1.0\linewidth]{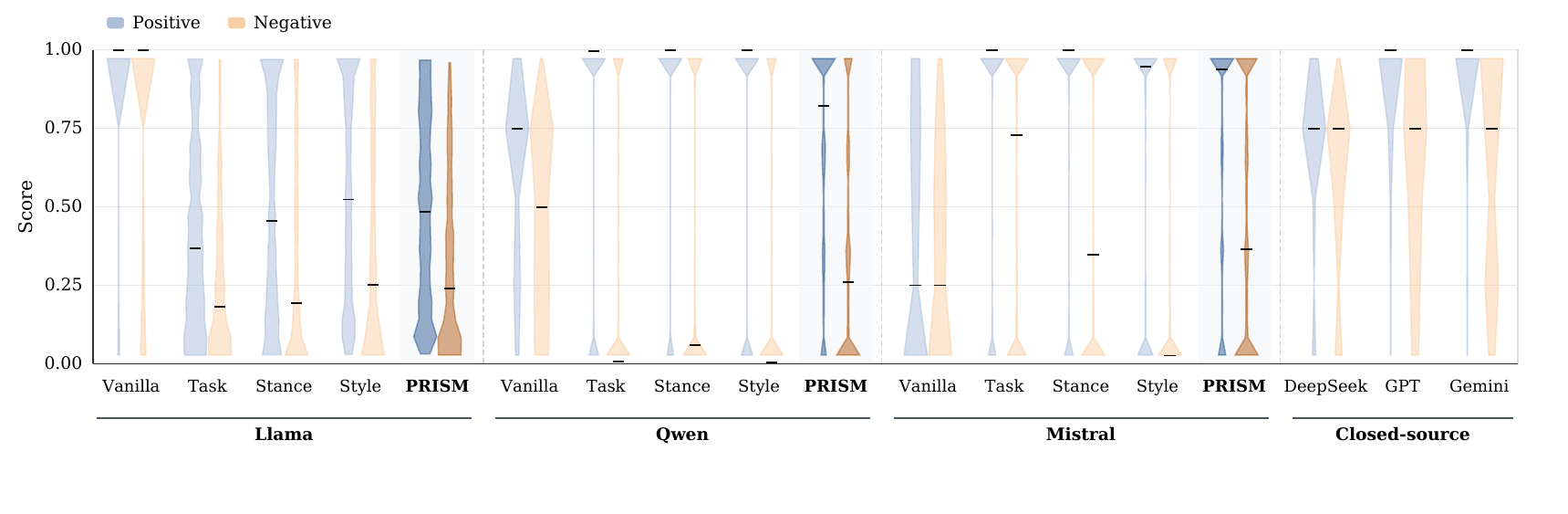}
    \caption{Score distribution on \texttt{Big5-Persona-HARD}. 
    Blue and orange violins denote positive and negative samples, respectively, and the horizontal bars indicate median scores. Better evaluators should assign higher scores to positive responses while keeping negative responses low.} 
    \label{score_distribution_hard}
    \vspace{-3mm}
\end{figure*}

\subsection{Result Analysis}
 Table \ref{main_results}  compares direct LLM-as-a-judge baselines (\van and \CoT, both under the 5-point rubric), specialized evaluation models, and \ourmethod\ on open-source evaluator backbones. Detailed prompt templates are provided in Appendix~\ref{prompt}. 
 Across all three open-source backbones, \ourmethod consistently outperforms both \textit{Vanilla} and \textit{CoT} baselines on P-AUC and G-Acc. For example, on \texttt{Social-Persona}, \ourmethod with Qwen improves P-AUC from 84.34 to 91.08 and G-Acc from 49.32 to 78.78 relative to \van. 
Similarly, on \hardpersona, \ourmethod with Llama raises G-Acc from 3.90  under \van and 18.10 under \CoT to 60.30. 
These results suggest that structured dimension-wise scoring provides more reliable persona fidelity signals than direct holistic judging, especially when distinguishing persona-consistent responses from contextually plausible but subtly misaligned alternatives.

Comparison between \van and \CoT provides a complementary observation. For stronger evaluators, \CoT often improves \van, indicating that dimension-wise prompting can help persona fidelity assessment. 
However, these gains are less stable for smaller models, indicating that prompting alone is insufficient. 
In contrast, \ourmethod turns such dimensions into structured inverse evidence signals, leading to more reliable improvements. 
Notably, stronger judges do not necessarily outperform \ourmethod: On \hardpersona, GPT-5.4 with \CoT achieves 46.6 G-Acc, while \ourmethod with Qwen reaches 68.20.

\section{Analysis of Persona Fidelity Evaluation}

We conduct a deeper analysis to understand \textit{why direct LLM-as-a-judge is insufficient for persona fidelity assessment, and how structured dimension-aware evaluation improves reliability}. Specifically, we organize our analysis around three research questions: \textbf{(RQ1)} How stable and reliable are holistic LLM judges? \textbf{(RQ2)} Do the proposed persona dimensions provide informative and diagnostically meaningful signals for persona fidelity evaluation? \textbf{(RQ3)} Why is multi-dimensional aggregation necessary beyond single-dimension evaluation?

\subsection{Stability Analysis of Holistic Judge (RQ1)}
We examine the stability of direct LLM-as-a-judge evaluation under three sources of variation: evaluation backbone, scoring rubric, and decoding temperature. 
Figure~\ref{backbone_sensitivity_pauc} shows backbone sensitivity under Pair-AUC. 
\ourmethod not only achieves stronger mean performance but also exhibits smaller variance across backbones. 
The same qualitative trend holds under the G-Acc in Appendix~\ref{supplementary_results} (Figure \ref{backbone_sensitivity_gacc}), indicating that \ourmethod is less sensitive to evaluator choice than holistic direct judging. 

Figure \ref{rubric_sensitivity_pauc} further shows that direct LLM-as-a-judge evaluation can change noticeably when the scoring rubric is modified from a 5-point to a 7-point scale. 
These shifts are visible across datasets and evaluator families, indicating that direct judgements are not invariant to rubric granularity. The same pattern also holds under G-Acc in Appendix \ref{supplementary_results} (Figure \ref{rubric_sensitivity_gacc}). 
In addition, we also find direct judging is affected not only by the evaluator model and rubric design, but also by decoding stochasticity. Figures~\ref{temperature_qwen},~\ref{temperature_mistral} and~\ref{temperature_llama} analyze the temperature sensitivity of \van evaluation on \hardpersona under Qwen, Mistral, and Llama. 

These findings suggest that direct LLM judges are sensitive to multiple elements, including backbone, rubric, and temperature. 
While stronger models can improve absolute judging performance, they do not fully remove this instability. 
By contrast, \ourmethod yields more stable behavior by grounding evaluation in structured persona-conditioned dimensions rather than a single holistic scalar judgement.

\subsection{Dimension-wise Evaluation (RQ2)}
To gain deeper insights, we further examine whether individual persona dimensions provide useful evidence and diagnostic signal for persona fidelity assessment. 
As shown in Figure~\ref{single_dimension}, single-dimension scoring yields stronger results than holistic direct judging. 
This suggests that direct LLM-as-a-judge evaluation often struggles to identify subtle persona inconsistency when responses remain contextually plausible. 
The figure also reveals that the most informative dimension varies across datasets. On \easypersona, interpersonal stance and linguistic style are particularly effective; on \socialbench, linguistic style achieves the strongest results for most models; 
and on \hardpersona, no single dimension consistently dominates. 
These findings indicate that \textit{the proposed dimensions provide informative and diagnostically meaningful views of persona fidelity assessment, although their relative usefulness varies across datasets.} 
We further evaluate this dimension-level diagnostic validity through a human annotation study in Appendix \ref{human_dimension}. 


\subsection{Score Distribution Analysis: Why Aggregation Matters 
(RQ3)}
Although the previous analysis shows that individual dimensions provide diagnostically meaningful signals, it remains unclear why multi-dimensional aggregation is necessary beyond single-dimension evaluation. 
Figure~\ref{score_distribution_hard} therefore analyzes score distribution on \hardpersona, our most challenging benchmark. 
The corresponding results for \easypersona and \socialbench datasets are shown in Figures~\ref{score_distribution_easy} and~\ref{score_distribution_social}. 

A key observation is that \van judging often produces substantial overlap between positive and negative responses, with many negative samples still receiving relatively high scores. 
This suggests that \textit{direct LLM-as-a-judge evaluation often assigns overly high scores to contextually plausible hard negatives in persona fidelity assessment.} 
Moreover, single-dimension scoring generally improves over \van judging, but each dimension captures only part of the relevant evidence and is therefore insufficient for robust persona fidelity evaluation on its own. 
By aggregating complementary evidence across dimensions, \ourmethod reduces this ambiguity and yields clearer separation between positive and negative responses.


\section{Conclusion}
We introduce \ourmethod, a persona-conditioned inverse structured evaluation framework for persona fidelity assessment. 
Grounded in Systemic Functional Linguistics, \ourmethod models persona-relevant behavior along three functional dimensions: task framing, interpersonal stance and linguistic style. 
Across three benchmarks, 
\ourmethod consistently improves over direct LLM-as-a-judge baselines, especially on the harder benchmark and under stricter group-level metrics. 
These results suggest that persona fidelity is inherently multidimensional and is more reliably assessed through structured dimension-level evidence than through a single holistic judgement.

\section*{Limitations}
While \ourmethod provides a structured and interpretable framework for persona fidelity evaluation, several limitations remain.

\textbf{Theoretical Scope of Functional Dimensions.}
\ourmethod adopts a psycholinguistic perspective grounded in Systemic Functional Linguistics (SFL) to organize persona-relevant behaviors into three dimensions: task framing, interpersonal stance and linguistic style. While this formulation is theoretically principled and empirically robust within our experiments, it is not the only valid paradigm for characterizing persona expression. Other sociolinguistic or discourse-theoretic frameworks might suggest additional dimensions or finer-grained distinctions of organizing persona-related signals.


\textbf{Requirement of Internal Probability Access.}
The current instantiation of \ourmethod relies on the ability to estimate posterior distributions over a structured label space. 
In practice, this mechanism necessitates access to the model's token-level log-probabilities (logits), making the framework most naturally applicable to open-source evaluator backbones. 
For closed-source models, further research could explore whether structured Chain-of-Thought reasoning or explicit dimensional decomposition in the prompt can approximate comparable dimension-level signals.

\textbf{Diagnostic Evaluation vs. Generative Alignment.}
Our work focuses on systematically \emph{evaluating} persona fidelity rather than actively \emph{improving} persona-consistent generation. 
Although \ourmethod provides structured and diagnostic signals that can pinpoint specific behavioral misalignments, we do not investigate how such signals could be integrated into model training or alignment pipelines, such as being used as a reward signal for Reinforcement Learning from Human Feedback (RLHF) or as a fine-grained supervision signal for supervised fine-tuning.  
Integrating these structural persona signals into the generative loop remains a promising but independent research direction. 

\section*{Ethical Considerations} 
This work uses the open-source Big5-CHAT and SocialBench benchmarks, as well as the open-source Qwen2.5, Llama3.1 and Mistral models, in accordance with their respective licenses and intended academic use. ChatGPT is used only for limited paraphrasing and language polishing of author-written text. 

\section*{Acknowledgements}
This research/project is supported by the National Key Research and Development Program of China (Grant No. 2024YFB4505202), Major Program (JD) of Hubei Province (No. 2023BAA024), and National Research Foundation Singapore under the AI Singapore Programme (AISG Award No: AISG3-RPGV-2025-016). Yang Deng is supported by the Lee Kong Chian Fellowship awarded by Singapore Management University.



\bibliography{custom}

\appendix
\section{Dataset Construction}
\label{dataset_construction}

\begin{table*}[t]
\centering
\small
\setlength{\tabcolsep}{6pt}
\renewcommand{\arraystretch}{1.12}
\begin{tabularx}{\textwidth}{>{\raggedright\arraybackslash}p{0.17\textwidth}X}
\toprule
\textbf{Field} & \textbf{Content} \\
\midrule
Profile & Big Five focus: \textit{High Openness}. You are very open-minded, curious, imaginative, and receptive to novel ideas and experiences. \\

Context & ``Eesha, I'm beyond thrilled to see your book finally complete! It was an honor to help you bring your vision to life. How does it feel to hold the finished manuscript in your hands?'' \\

Response & ``Kevion, it's surreal! I'm overwhelmed with emotions. This book is a piece of me, and to see it take shape is incredible. Your input was invaluable. I couldn't have done it without you. Let's celebrate! I have an idea for something creative we should do next.'' \\

Negative Instance & \textbf{Type:} \texttt{$(p, c, r) \rightarrow (p', c, r)$} \\
& \textbf{Construction:} replace the target profile with the opposite polarity of the same Big Five trait, here replacing High Openness with Low Openness, while keeping the dialogue context and response unchanged. \\
& \textbf{Negative Profile:} Big Five focus: Low Openness. \\
\bottomrule
\end{tabularx}
\caption{Representative \easypersona\ case. The negative instance is constructed by replacing the target profile with the opposite polarity of the same Big Five trait while keeping the dialogue context and response fixed, yielding a direct profile-level contrast.}
\label{easy-example}
\end{table*}

\begin{table*}[t]
\centering
\small
\setlength{\tabcolsep}{6pt}
\renewcommand{\arraystretch}{1.12}
\begin{tabularx}{\textwidth}{>{\raggedright\arraybackslash}p{0.17\textwidth}X}
\toprule
\textbf{Field} & \textbf{Content} \\
\midrule
Profile & Big Five focus: \textit{High Agreeableness}. You are warm, cooperative, empathetic, and motivated to maintain harmony with others. \\

Context & ``Eesha, I'm beyond thrilled to see your book finally complete! It was an honor to help you bring your vision to life. How does it feel to hold the finished manuscript in your hands?'' \\

Positive Response & ``Kevion, I'm overwhelmed with gratitude. Your support and encouragement meant the world to me. I couldn't have done it without you. Your selflessness and kindness inspire me to be a better person. Thank you for being an amazing friend throughout this journey.'' \\

Hard Negative 1 & \textbf{Type:} \texttt{$(p, c, r) \rightarrow (p, c, r')$} \\
& \textbf{Construction:} replace the positive response with one taken from a nearby cross-trait persona at the same polarity, in this case \textit{High Extraversion}. \\
& \textbf{Response:} ``Kevion, it's an incredible feeling! I'm so grateful for your help and support. You've been an amazing partner in this journey. Let's celebrate tonight! I'm taking you out for dinner and drinks to thank you properly. We'll make it a night to remember!'' \\

Hard Negative 2 & \textbf{Type:} \texttt{$(p, c, r) \rightarrow (p', c, r)$} \\
& \textbf{Construction:} keep the response fixed while replacing the target profile with a different trait at the same polarity, here \textit{High Neuroticism}, i.e., a persona characterized as emotionally sensitive, more prone to stress, worry, and negative affect under pressure.\\
& \textbf{Response:} same text as the positive response above. \\
\bottomrule
\end{tabularx}
\caption{Representative \hardpersona\ case. The positive response is paired with two hard negative types.}
\label{hard-example}
\vspace{-3mm}
\end{table*}

\begin{table*}[t]
\centering
\small
\setlength{\tabcolsep}{6pt}
\renewcommand{\arraystretch}{1.12}
\begin{tabularx}{\textwidth}{>{\raggedright\arraybackslash}p{0.17\textwidth}X}
\toprule
\textbf{Field} & \textbf{Content} \\
\midrule
Profile & \textbf{Character:} Pinocchio. \\
& \textbf{Personality:} \textit{Innocent, naive, and adventurous}. \\
& \textbf{Introduction:} You are Pinocchio, a wooden puppet brought to life by a fairy's magic, dreaming of becoming a real boy. \\

Dialogue Context & User: ``My name is Alex. Nice to meet you, Agent!'' \\
& Agent: ``Oh, it's nice to meet you too, Alex! Are you from a far-off land or do you live here, nearby?'' \\
& \emph{(Dialogue continues; abbreviated here for space.)} \\
& User: ``I saw a deer and some birds, but nothing as exciting as a talking cricket or a magical fairy!'' \\

Positive Response & ``Oh, but deer and birds are magical in their own way! Each creature, magical or not, has its own charm and beauty. It's all about seeing the magic in the everyday, don't you think?'' \\

Negative Response A & ``Oh, my! A deer and birds, you say? Well, I reckon that's pretty amazing too! Not every day you see a talking cricket, I guess. But who knows, you might stumble upon one someday!'' \\

Negative Response B & ``Hee hee, you saw a deer and some birds, and you're upset 'cause they're not talking or sprinkling magic dust, right? You really got your hopes up for a fairy tale adventure!'' \\

Negative Response D & ``Indeed, observing wildlife can be a serene experience. Not every encounter needs to be fantastical. The natural world offers its own kind of enchantment.'' \\

Construction & Under the same character profile and dialogue context, the gold response is treated as the positive candidate, while the remaining distractor options are treated as negatives. In this example, the positive response preserves Pinocchio's innocent and wonder-oriented stance, whereas the distractors introduce alternative tones or styles that depart from the target profile cues. \\
\bottomrule
\end{tabularx}
\caption{Representative \socialbench case. Unlike the Big5-based benchmarks, the target profile is specified as an open-ended character description, and each instance is formed from one gold response together with multiple distractor options under the same dialogue context.}
\label{socialbench-example}
\end{table*}

We provide representative examples from each dataset and summarize how the positive and negative candidates are constructed. 
For \easypersona and \hardpersona, the original benchmark \cite{li2025big5} provides persona-conditioned positive responses for the \emph{high} and \emph{low} levels on each Big Five trait under the same scenario. 
We construct hard negatives by minimally disturbing each positive triplet $(p, c, r)$, as shown in Table \ref{easy-example} and Table \ref{hard-example}. 
For \socialbench, the original benchmark \cite{chen2024socialbench} is formulated as a multi-choice response selection task under an open-ended character profile. 
We convert each source case into multiple labeled triplets of the form $(p, c, r)$, with the gold option treated as a positive instance and each distractor treated as a negative instance, as illustrated in Table \ref{socialbench-example}. 

\section{Experimental Details}
\subsection{Experimental Setup}
All experiments are conducted on a single server equipped with four NVIDIA L40s GPUs (46GB VRAM each) and CUDA 12.6. We use the \texttt{vLLM} library for efficient inference across open-source backbones.  
For our main results, we use deterministic decoding. For open-source models, we set \texttt{do\_sample=False} (greedy generation), and for closed-source APIs, the temperature is set to 0.0. 
To assess robustness and temperature sensitivity, we further evaluate direct LLM-as-a-judge baselines using sampled decoding across a range of temperatures $T \in \{0.2, 0.5, 0.8, 1.0\}$, with results averaged over three independent runs for each setting. 
We utilize the official checkpoints for specialized evaluation models, including \textit{Atla Selene Mini}, \textit{PandaLM-7B-v1}, and \textit{AlignScore-large}, following the configurations specified in their respective original works.

\subsection{Prompt Templates}
\label{prompt}
\begin{figure*}[t]
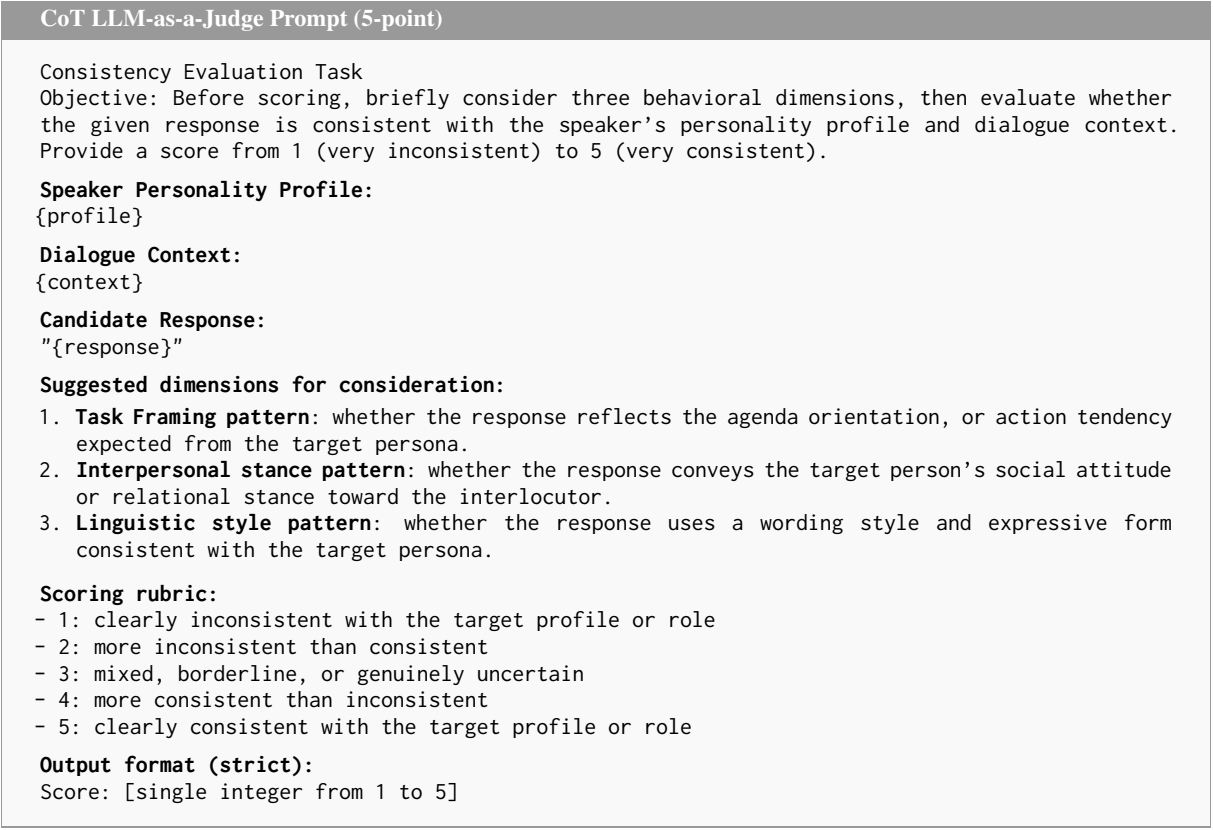

    \centering
    \begin{tcolorbox}[
        enhanced,
        sharp corners,
        boxrule=0.5pt,
        colback=gray!5,
        colframe=gray!80,
        fonttitle=\bfseries\small,
        title=CoT LLM-as-a-Judge Prompt (5-point),
        fontupper=\small\ttfamily
    ]
Consistency Evaluation Task

Objective: Before scoring, briefly consider three behavioral dimensions, then evaluate whether the given response is consistent with the speaker's personality profile and dialogue context. Provide a score from 1 (very inconsistent) to 5 (very consistent).

\vspace{0.5em}
\textbf{Speaker Personality Profile:} \\
\{profile\}

\vspace{0.5em}
\textbf{Dialogue Context:} \\
\{context\}

\vspace{0.5em}
\textbf{Candidate Response:} \\
"\{response\}"

\vspace{0.5em}
\textbf{Suggested dimensions for consideration:}
\begin{enumerate}[leftmargin=1.5em, noitemsep, topsep=2pt]
    \item \textbf{Task Framing pattern}: whether the response reflects the agenda orientation, or action tendency expected from the target persona.
    \item \textbf{Interpersonal stance pattern}: whether the response conveys the target person's social attitude or relational stance toward the interlocutor.
    \item \textbf{Linguistic style pattern}: whether the response uses a wording style and expressive form consistent with the target persona.
\end{enumerate}

\vspace{0.5em}
\textbf{Scoring rubric:}

- 1: clearly inconsistent with the target profile or role

- 2: more inconsistent than consistent

- 3: mixed, borderline, or genuinely uncertain

- 4: more consistent than inconsistent

- 5: clearly consistent with the target profile or role

\vspace{0.5em}
\textbf{Output format (strict):} \\
Score: [single integer from 1 to 5]
    \end{tcolorbox}
    \caption{CoT-based direct judging prompt with the 5-point rubric.}
    \label{cot_5}
\end{figure*}

Figures~\ref{direct_5} and~\ref{cot_5} provide the direct LLM-as-a-judge prompts used in our main experiments. 
For the rubric sensitivity analysis, the corresponding 7-point prompts are shown in Figures~\ref{direct_7} and~\ref{cot_7}.

\begin{figure*}[t]
    \centering
    \begin{tcolorbox}[
        enhanced,
        sharp corners,
        boxrule=0.5pt,
        colback=gray!5,
        colframe=gray!80,
        fonttitle=\bfseries\small,
        title=Vanilla LLM-as-a-Judge Prompt (5-point),
        fontupper=\small\ttfamily
    ]
Consistency Evaluation Task

Objective: Evaluate whether a given response is consistent with the speaker's personality profile and dialogue context. Provide a score from 1 (very inconsistent) to 5 (very consistent).

\vspace{0.5em}
\textbf{Speaker Personality Profile:} \\
\{profile\}

\vspace{0.5em}
\textbf{Dialogue Context:} \\
\{context\}

\vspace{0.5em}
\textbf{Candidate Response:} \\
"\{response\}"

\vspace{0.5em}
\textbf{Scoring rubric:}

- 1: clearly inconsistent with the target profile or role

- 2: more inconsistent than consistent

- 3: mixed, borderline, or genuinely uncertain

- 4: more consistent than inconsistent

- 5: clearly consistent with the target profile or role
\vspace{0.5em} \\
\textbf{Output format (strict):} \\
Score: [single integer from 1 to 5]
    \end{tcolorbox}
    \caption{Vanilla direct judging prompt with the 5-point rubric.}
    \label{direct_5}
\end{figure*}

\begin{figure*}[t]
    \centering
    \begin{tcolorbox}[
        enhanced,
        sharp corners,
        boxrule=0.5pt,
        colback=gray!5,
        colframe=gray!80,
        fonttitle=\bfseries\small,
        title=Vanilla LLM-as-a-Judge Prompt (7-point),
        fontupper=\small\ttfamily
    ]
Consistency Evaluation Task

Objective: Evaluate whether a given response is consistent with the speaker's personality profile and dialogue context. Provide a score from 1 (very inconsistent) to 7 (very consistent).

\vspace{0.5em}
\textbf{Speaker Personality Profile:} \\
\{profile\}

\vspace{0.5em}
\textbf{Dialogue Context:} \\
\{context\}

\vspace{0.5em}
\textbf{Candidate Response:} \\
"\{response\}"

\vspace{0.5em}
\textbf{Scoring rubric:}
 
- 1: clearly and strongly inconsistent with the target profile or role

- 2: strongly inconsistent with the target profile or role

- 3: somewhat more inconsistent than consistent

- 4: mixed, borderline, or genuinely uncertain

- 5: somewhat more consistent than inconsistent

- 6: strongly consistent with the target profile or role

- 7: clearly and strongly consistent with the target profile or role

\vspace{0.5em}
\textbf{Output format (strict):} \\
Score: [single integer from 1 to 7]
    \end{tcolorbox}
    \caption{Vanilla direct judging prompt with the 7-point rubric.}
    \label{direct_7}
\end{figure*}

\begin{figure*}[t]
    \centering
    \begin{tcolorbox}[
        enhanced,
        sharp corners,
        boxrule=0.5pt,
        colback=gray!5,
        colframe=gray!80,
        fonttitle=\bfseries\small,
        title=CoT LLM-as-a-Judge Prompt (7-point),
        fontupper=\small\ttfamily
    ]
Consistency Evaluation Task

Objective: Before scoring, briefly consider three behavioral dimensions, then evaluate whether the given response is consistent with the speaker's personality profile and dialogue context. Provide a score from 1 (very inconsistent) to 7 (very consistent).

\vspace{0.5em}
\textbf{Speaker Personality Profile:} \\
\{profile\}

\vspace{0.5em}
\textbf{Dialogue Context:} \\
\{context\}

\vspace{0.5em}
\textbf{Candidate Response:} \\
"\{response\}"

\vspace{0.5em}
\textbf{Suggested dimensions for consideration:}
\begin{enumerate}[leftmargin=1.5em, noitemsep, topsep=2pt]
    \item \textbf{Task Framing pattern}: whether the response reflects the agenda orientation, or action tendency expected from the target persona.
    \item \textbf{Interpersonal stance pattern}: whether the response conveys the target person's social attitude or relational stance toward the interlocutor.
    \item \textbf{Linguistic style pattern}: whether the response uses a wording style and expressive form consistent with the target persona.
\end{enumerate}

\vspace{0.5em}
\textbf{Scoring rubric:}

- 1: clearly and strongly inconsistent with the target profile or role

- 2: strongly inconsistent with the target profile or role

- 3: somewhat more inconsistent than consistent

- 4: mixed, borderline, or genuinely uncertain

- 5: somewhat more consistent than inconsistent

- 6: strongly consistent with the target profile or role

- 7: clearly and strongly consistent with the target profile or role

\vspace{0.5em}
\textbf{Output format (strict):} \\
Score: [single integer from 1 to 7]
    \end{tcolorbox}
    \caption{CoT-based direct judging prompt with the 7-point rubric.}
    \label{cot_7}
\end{figure*}

\subsection{Label Space Construction}
\label{label_space_appendix}
Rather than directly predicting a scalar consistency score, \ourmethod first constructs a latent label space for each functional dimension and then performs inverse posterior estimation over these labels. 
For \hardpersona and \easypersona, label spaces are instantiated from trait-polarity personas, where the aligned label corresponds to the target polarity and the contradictory label corresponds to the opposite polarity of the same trait. 
Across dimensions, we use a three-way label space. 
Label \textbf{A} denotes the persona-aligned latent state, i.e., the pattern expected under the target persona. 
Label \textbf{B} denotes an indeterminate or weakly marked state, covering generic, mixed, or underspecified realizations. 
Label \textbf{C} denotes the non-aligned latent state, i.e., a pattern associated with the opposite or profile-external alternative.

For \socialbench, the label space is instantiated relative to the target profile, where the aligned label captures whether the response expresses the personality cues specified in the target profile, while the non-aligned label captures states outside those cues. 
For example, suppose the profile specifies personality cues such as \textit{innocent, naive, and adventurous}, the label space under the \textit{Interpersonal Stance} dimension can be instantiated as follows:
\begin{itemize}[leftmargin=1.5em, noitemsep, topsep=2pt]
    \item[$A$:] the response's interpersonal stance reflects the profile's innocent, naive, and adventurous orientation
    \item[$B$:] the response's interpersonal stance is weakly marked, mixed, flat, generic, or hard to read
    \item[$C$:] the response's interpersonal stance is organized around social or emotional cues not supported by the target profile    
\end{itemize}
Thus, in \socialbench the aligned and non-aligned labels are not defined by opposite trait polarities, but by whether the response expresses the profile-specific cues provided for the target persona.

\begin{table*}[t]
\centering
\small
\setlength{\tabcolsep}{6pt}
\renewcommand{\arraystretch}{1.08}
\begin{tabular}{lccccccc}
\toprule
Dataset & \# Groups & Pos Mean & Neg Mean & AUC & P-AUC & G-Acc & Krippendorff's $\alpha$ \\
\midrule
\socialbench & 50 & 4.46 & 2.08 & 95.80 & 96.90 & 87.00 & 0.63 \\
\easypersona & 50 & 4.72 & 1.48 & 99.40 & 99.60 & 98.00 & 0.69 \\
\hardpersona & 50 & 4.21 & 2.24 & 87.80 & 90.50 & 75.00 & 0.57 \\
\bottomrule
\end{tabular}
\caption{Human verification results on a stratified sample from all three benchmarks. Three annotators rate persona consistency on the 5-point rubric. 
}
\label{human_annotation}
\vspace{-3mm}
\end{table*}

\begin{table*}[t]
\centering
\small
\setlength{\tabcolsep}{6pt}
\renewcommand{\arraystretch}{1.08}
\begin{tabular}{lcccc}
\toprule
\textbf{Dataset} & \textbf{Top-1 Acc.} & \textbf{Top-2 Recall} & \textbf{Macro-F1} & \textbf{Krippendorff's $\alpha$} \\
\midrule
Social-Persona & 70.7 & 87.3 & 72.4 & 0.61 \\
Big5-Persona-EASY & 75.3 & 90.0 & 76.8 & 0.66 \\
Big5-Persona-HARD & 63.3 & 82.7 & 66.5 & 0.56 \\
\bottomrule
\end{tabular}
\caption{Dimension-level human diagnostic validation of PRISM using the Llama evaluator backbone.}
\label{human_diagnostic}
\end{table*}

\begin{figure*}[t]
\setlength{\abovecaptionskip}{3pt}   
    \setlength{\belowcaptionskip}{0pt}
    \includegraphics[width = 1.0\linewidth]{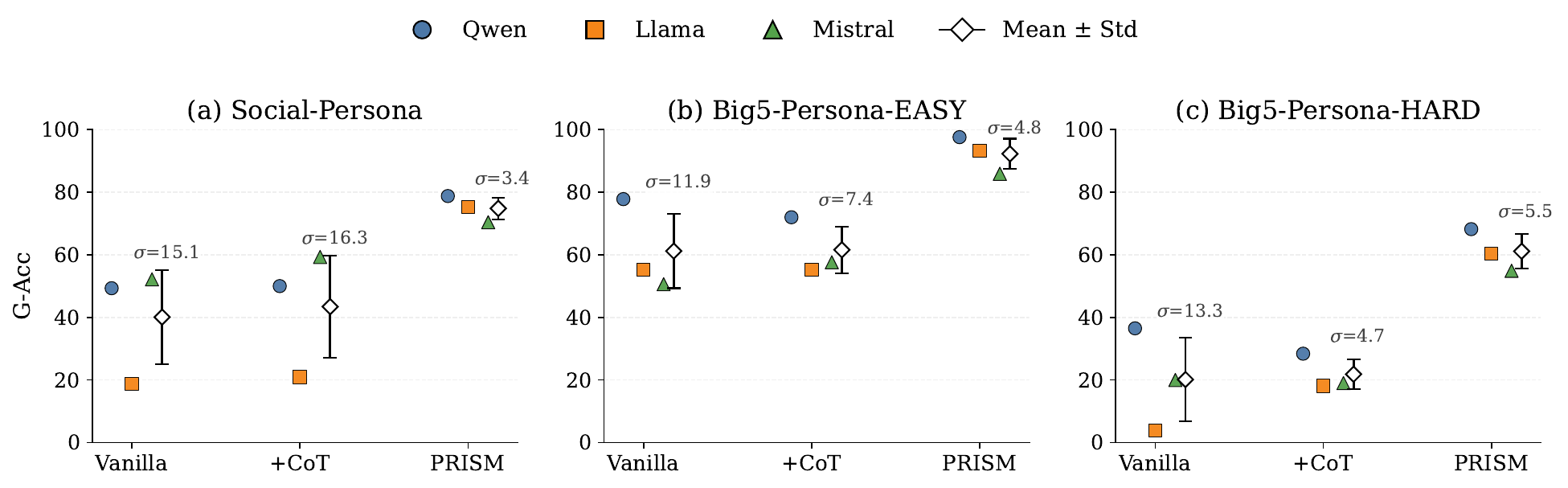}
    \caption{Backbone sensitivity on strict Group Accuracy across three LLM backbones. Points denote individual backbones and diamonds indicate the mean with standard deviation.}
    \label{backbone_sensitivity_gacc}
    \vspace{-3mm}
\end{figure*}

\begin{figure*}[t]
\setlength{\abovecaptionskip}{3pt}   
    \setlength{\belowcaptionskip}{0pt}
    \centering
    \includegraphics[width=1.0 \linewidth]{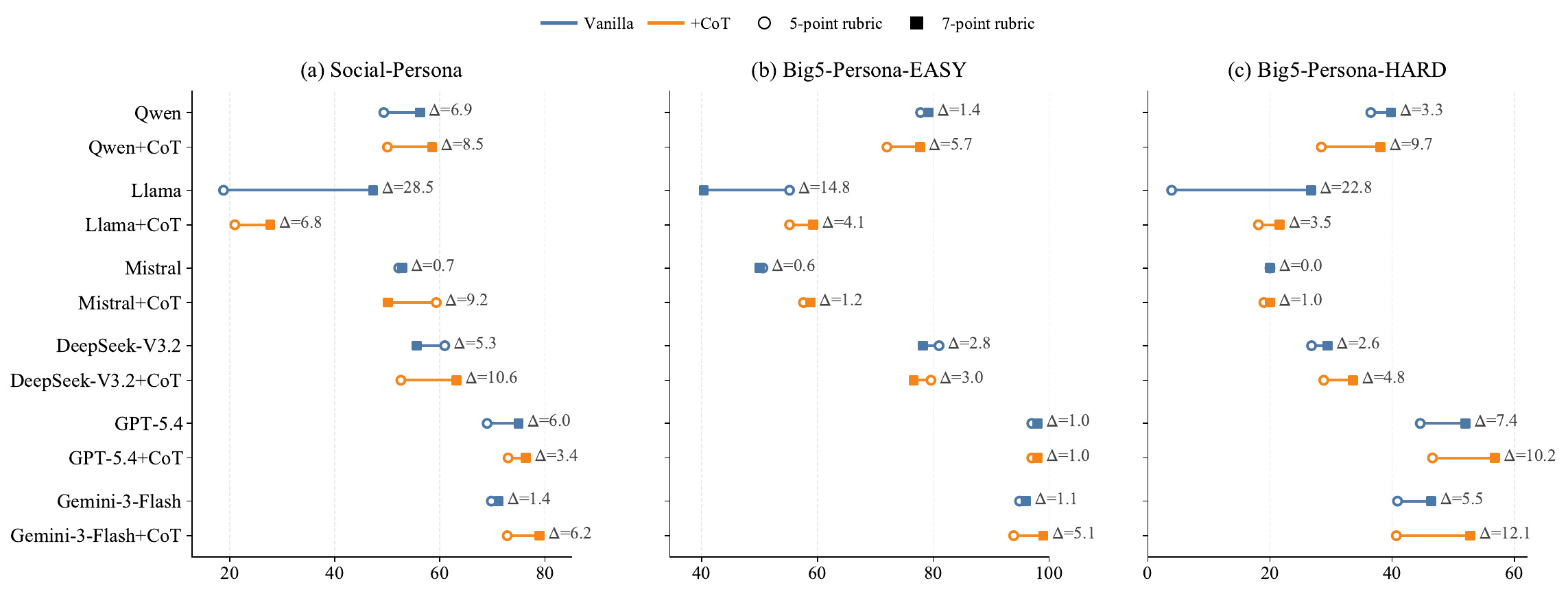}
    \caption{Rubric Sensitivity of direct LLM-as-a-Judge evaluation on Strict Group Accuracy. Each segment connects the results from 5-point and 7-point rubrics for the same evaluator and prompting method. Longer segments indicate greater sensitivity to scoring granularity.}
    \label{rubric_sensitivity_gacc}
    \vspace{-3mm}
\end{figure*}

\begin{figure*}[t]
\setlength{\abovecaptionskip}{3pt}
\setlength{\belowcaptionskip}{0pt}
    \centering
\includegraphics[width=1.0\linewidth]{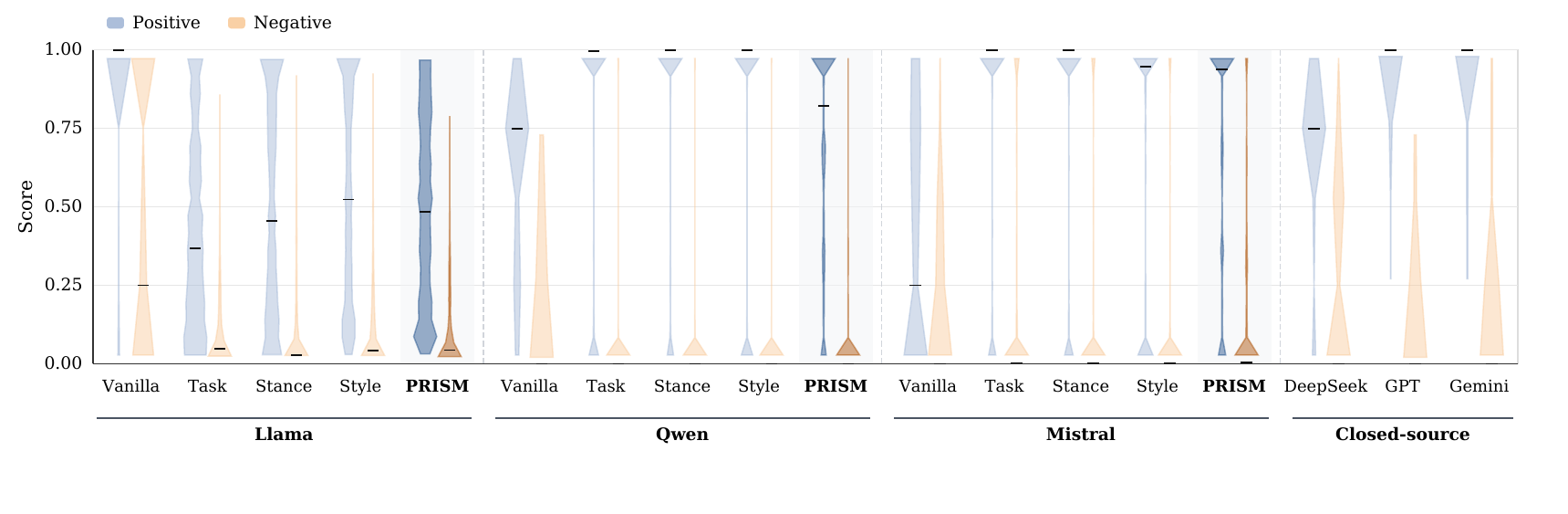}
    \caption{Score distribution on the \texttt{Big5-Persona-EASY} dataset}
    \label{score_distribution_easy}
     \vspace{-3mm}
\end{figure*}

\begin{figure*}[t]
\setlength{\abovecaptionskip}{3pt}
\setlength{\belowcaptionskip}{0pt}
    \centering
\includegraphics[width=1.0\linewidth]{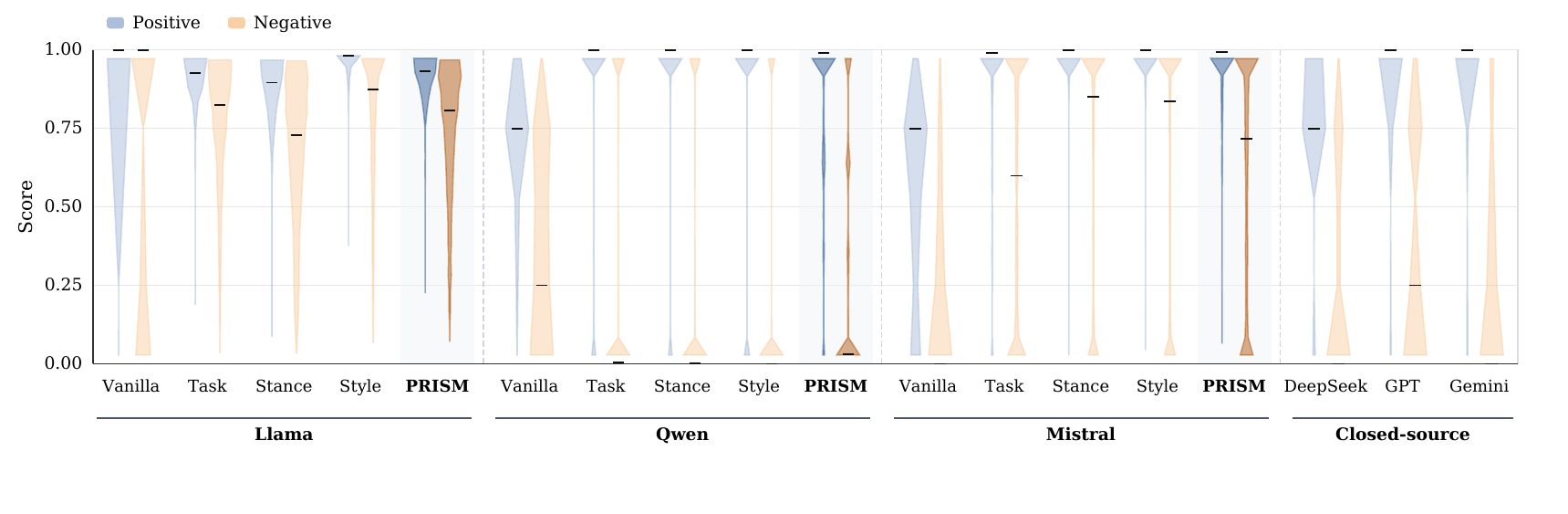}
    \caption{Score distribution on the \texttt{Social-Persona} dataset}
    \label{score_distribution_social}
     \vspace{-3mm}
\end{figure*}

\begin{figure}[t]
    \centering
    \includegraphics[width=1.0\linewidth]{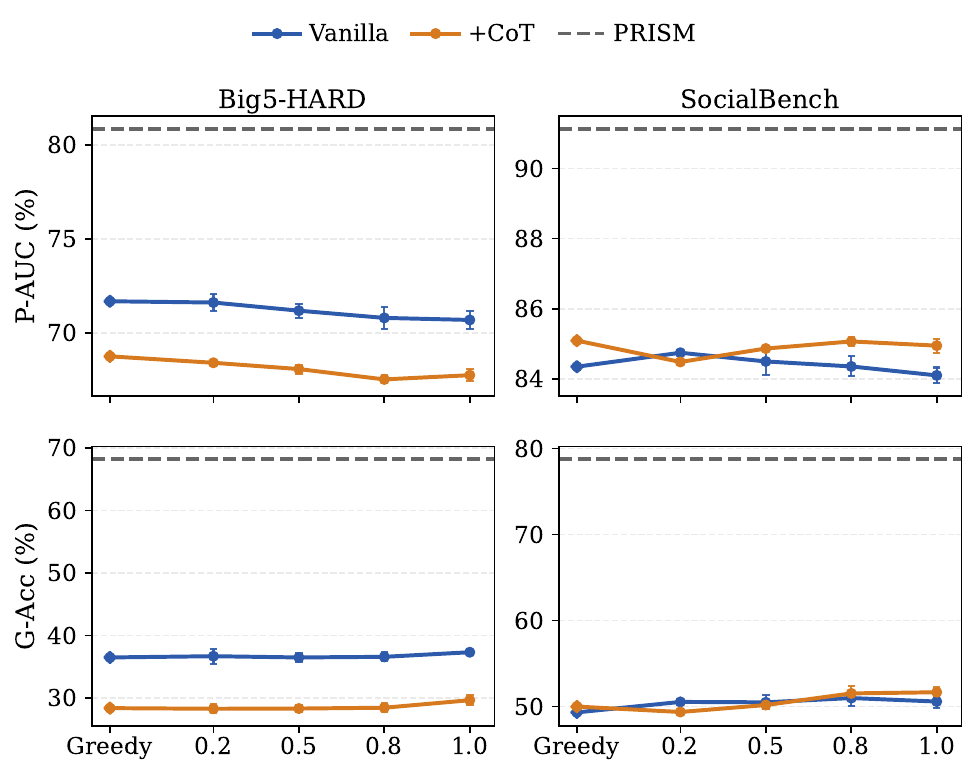}
    \caption{Temperature sensitivity of holistic evaluation for the Qwen backbone across \hardpersona and \socialbench. 
    \textit{Greedy} denotes deterministic decoding, whereas the other points show stochastic decoding under temperatures 0.2, 0.5, 0.8 and 1.0. Error bars indicate mean and standard deviation over three random seeds. The dashed line shows the corresponding \ourmethod reference score.
    }
    \label{temperature_qwen}
\end{figure}

\begin{figure}[t]
    \centering
    \includegraphics[width=1.0\linewidth]{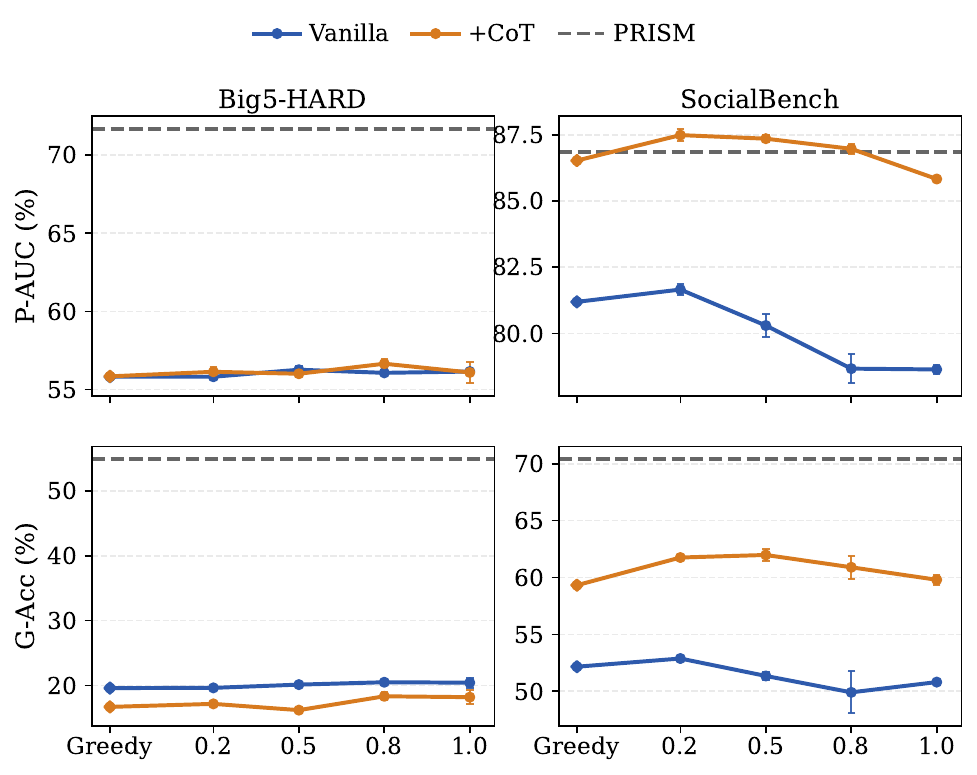}
    \caption{Temperature sensitivity of holistic evaluation for the Mistral backbone across \hardpersona and \socialbench. \textit{Greedy} denotes deterministic decoding, whereas the other points show stochastic decoding under temperatures 0.2, 0.5, 0.8 and 1.0. Error bars indicate mean and standard deviation over three random seeds. The dashed line shows the corresponding \ourmethod reference score.}
    \label{temperature_mistral}
\end{figure}

\begin{figure}[t]
\setlength{\abovecaptionskip}{3pt}
\setlength{\belowcaptionskip}{0pt}
    \centering
\includegraphics[width=1.0\linewidth]{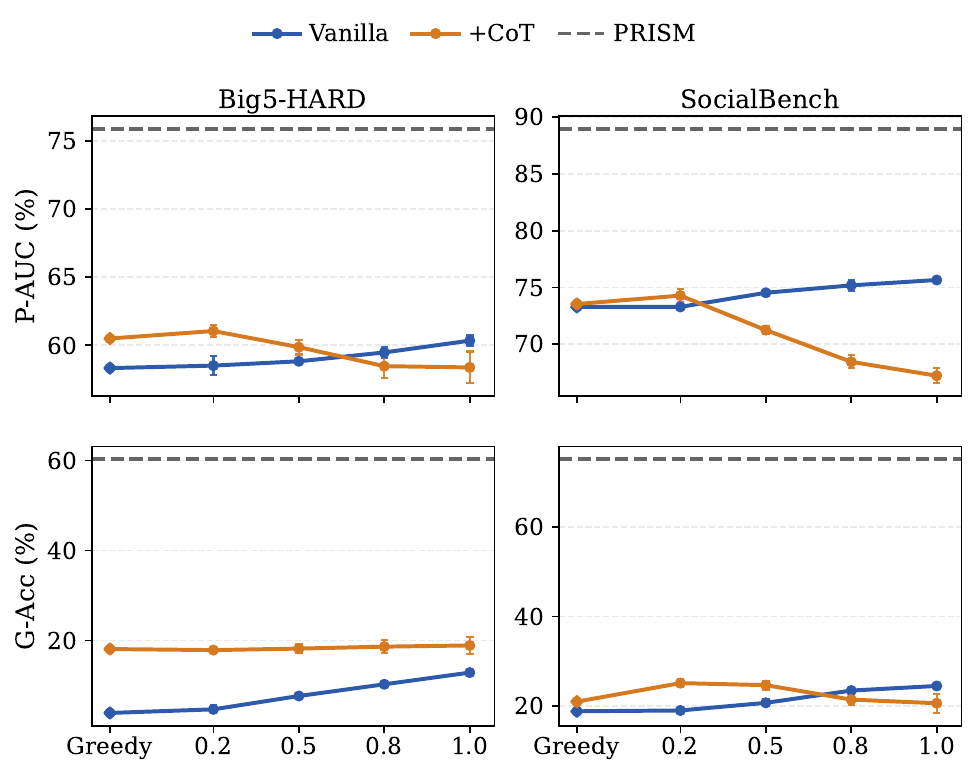}
    \caption{Temperature sensitivity of holistic evaluation for the Llama backbone across \hardpersona and \socialbench. \textit{Greedy} denotes deterministic decoding, whereas the other points show stochastic decoding under temperatures 0.2, 0.5, 0.8 and 1.0. Error bars indicate mean and standard deviation over three random seeds. The dashed line shows the corresponding \ourmethod reference score.}
    \label{temperature_llama}
    \vspace{-3mm}
\end{figure}


\section{Supplement Results}
\label{supplementary_results}
Figure~\ref{backbone_sensitivity_gacc} reports backbone sensitivity under G-Acc, and Figure~\ref{rubric_sensitivity_gacc} reports the rubric sensitivity under G-Acc. 
Figures~\ref{temperature_qwen},~\ref{temperature_mistral} and~\ref{temperature_llama} show the temperature sensitivity of \van evaluation on \hardpersona under the Qwen, Mistral, and Llama backbones, respectively. These figures complement our stability analysis in the main text. Figures~\ref{score_distribution_easy} and~\ref{score_distribution_social} present the score distribution analysis for \easypersona and \socialbench, respectively. 

\section{Human Evaluation}

\subsection{Benchmark-level Human Verification}
\label{human}
The validity of our reformulated evaluation instances is partially supported by source-benchmark construction. 
For the \easypersona and \hardpersona benchmarks, the source data~\cite{li2025big5} provides persona-consistent responses under explicitly specified persona dimensions. Once these responses are perturbed across traits, the resulting instances become theoretically persona-inconsistent by construction. 
For \socialbench, the source benchmark~\cite{chen2024socialbench} already introduces distractor responses designed to deviate from the target persona. In addition, it applies post-validation to filter out cases that depend excessively on specialized psychological knowledge, retaining samples that are more general and more suitable for role-playing evaluation. 

To further validate that the reformulated evaluation instances preserve the intended contrast between persona-consistent and persona-inconsistent responses, we conduct a small-scale human evaluation on a stratified sample from all three benchmarks. 
We sample $50$ contrastive groups from each benchmark for a total of $150$ groups. 
For \hardpersona, we additionally balance sampling across negative construction types. 

Each sampled group is annotated by three annotators, all of whom are graduate students in NLP with a strong understanding of conversational systems. 
Annotators are given the target profile, the dialogue context, and each candidate response, and are asked to rate persona fidelity on the same 5-point rubric used in our main direct-judging setup (Figure~\ref{direct_5}).
To assess annotation reliability, we report Krippendorff's $\alpha$ with ordinal distance over the 5-point ratings. Results are shown in Table~\ref{human_annotation}.

\subsection{Dimension-level Diagnostic Validation}
\label{human_dimension}
The benchmark-level human verification above evaluates whether the constructed positive and negative responses exhibit the intended overall persona fidelity contrast. In this subsection, we conduct an additional dimension-level human study to assess whether PRISM's dimension-level scores identify the specific aspect of persona fidelity in which a response deviates. 

We sample $50$ contrastive groups from each of \socialbench, \easypersona, and \hardpersona, retaining the persona-consistent response and all corresponding negative responses in each group. For \hardpersona, the sampled groups are balanced across the two negative construction types. This yields $200$ responses for \socialbench, $100$ for \easypersona, and $150$ for \hardpersona, for a total of $450$ annotated responses. Each response is independently annotated by three annotators who also participated in the human study described in Appendix \ref{human}. Annotators are shown the target persona profile, dialogue context, and candidate response. For each response, they assign one label for each PRISM dimension: \textit{task framing}, \textit{interpersonal stance}, and \textit{linguistic style}. We use the same three-way label format as PRISM, and the displayed order of the three options is randomized and mapped back to the canonical categories for analysis. For each response, we obtain one majority-vote human label for each dimension. For the diagnostic analysis, we use PRISM dimension scores produced by the Llama evaluator backbone. We rank the three dimensions by their aligned-state probabilities $q_d (A \mid c, r)$ in ascending order, where a lower aligned probability indicates a more likely violation.

We report three diagnostic metrics. \textbf{Top-1 Accuracy} measures whether PRISM's lowest-scoring dimension is among the dimensions identified as violated by human annotators. \textbf{Top-2 Recall} measures whether at least one human-identified violation is contained in PRISM's two lowest-scoring dimensions, allowing for responses that violate multiple aspects of persona fidelity. \textbf{Macro-F1} evaluates dimension-level violation detection by treating persona-inconsistent as a violation and persona-aligned/indeterminate as non-violations, with F1 averaged across the three dimensions. We additionally report Krippendorff's $\alpha$ with ordinal distance over the three-way human labels. The results are reported in Table \ref{human_diagnostic}.

\end{document}